\documentclass[final,5p,times,twocolumn]{elsarticle}

\usepackage{amssymb}
\usepackage{amsmath}
\usepackage{multirow}
\usepackage[table]{xcolor}
\usepackage{booktabs}

\usepackage{multirow}

\usepackage{pifont}
\newcommand{\cmark}{\ding{51}}

\journal{Neurocomputing}

\begin{document}

\begin{frontmatter}



\title{Beyond Global Scanning: Adaptive Visual State Space Modeling for Salient Object Detection in Optical Remote Sensing Images}

\affiliation[1]{
  organization={School of Computer Science and Technology, Hainan University},
  city={Haikou},
  postcode={570228},
  country={China}
}

\affiliation[2]{
  organization={School of Control Science and Engineering, Shandong University},
  city={Jinan},
  postcode={250061},
  country={China}
}

\fntext[†]{These authors contributed equally to this work.}
\cortext[cor1]{Corresponding author.}

\author[1]{Mengyu Ren\fnref{†}}
\author[1]{Yutong Li\fnref{†}}
\author[1]{Hua Li\corref{cor1}}
\author[1]{Chuhong Wang}
\author[2]{Runmin Cong}

\begin{abstract}
Salient object detection (SOD) in optical remote sensing images (ORSIs) faces numerous challenges, including significant variations in target scales and low contrast between targets and the background. Existing methods based on vision transformers (ViTs) and  convolutional neural networks (CNNs) architectures aim to leverage both global and local features, but the difficulty in effectively integrating these heterogeneous features limits their overall performance. To overcome these limitations, we propose an adaptive state space context network (ASCNet), which builds upon the state space model mechanism to simultaneously capture long-range dependencies and enhance regional feature representation. Specifically, we employ the visual state space encoder to extract multi-scale features. To further achieve deep guidance and enhancement of these features, we design a 
Multi-Level Context Module (MLCM), which module strengthens cross-layer interaction capabilities between features of different scales while enhancing the model’s structural perception,
allowing it to distinguish between foreground and background more effectively. Then, we design the Adaptive Patchwise Visual State Space (APVSS) block as the decoder of ASCNet, which integrates our proposed Dynamic Adaptive Granularity Scan (DAGS) and Granularity-aware Propagation Module (GPM).
It performs adaptive patch scanning on feature maps enhanced by local perception, thereby capturing rich local region information and enhancing state space model’s local modeling capability. Extensive experimental results demonstrate that the proposed model achieves state-of-the-art performance, validating its effectiveness and superiority.
\end{abstract}




\end{frontmatter}



\section{Introduction}
\label{sec1}
\begin{figure}[t]
  \centering
  \includegraphics[width=\columnwidth]{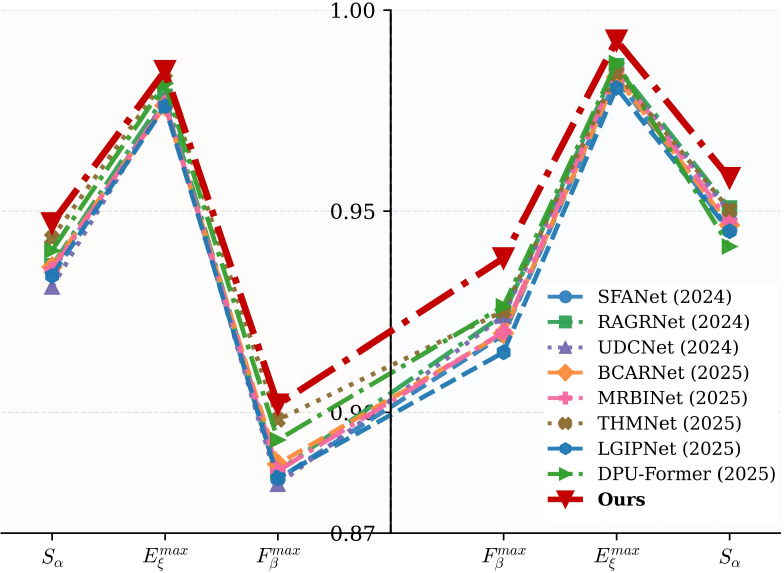}
  \caption{Overall performance comparison on the \textbf{EORSSD}(L) and \textbf{ORSSD}(R) datasets.
  Our method consistently outperforms recent SOTA approaches across multiple evaluation metrics,
  demonstrating stable and superior performance on both datasets.}
  \label{fig:fig1}
\end{figure}

Salient Object detection (SOD) is a fundamental task in the field of computer vision, aiming to identify and segment the most attention-grabbing objects or regions within a scene\cite{1_2015,1_2018}. 
As a key prerequisite visual task, research in SOD has been widely applied to areas such as object tracking\cite{OT_2015,OT_2019}, image segmentation\cite{seg_2019,seg_2020}, and underwater exploration\cite{li2025waterflow}, as well as many other fields\cite{QA_2019,qiu2025salient}.
Traditionally, SOD methods have primarily focused on natural scene images and have achieved remarkable results in that domain. In recent years, with the continuous advancement of satellite remote sensing technology, salient object detection in optical remote sensing images has increasingly attracted the attention of researchers. Compared with natural images, ORSIs pose numerous challenges for salient object detection, such as low target-background contrast, significant variations in target scale, and complex object structures. These factors make it difficult for many SOD methods that perform well on natural images to be directly transferred to remote sensing scenarios. As a result, researchers have developed specialized models tailored to the characteristics of remote sensing images, aiming to effectively detect salient objects in optical remote sensing imagery.

Traditional optical remote sensing image salient object detection (ORSI-SOD) methods\cite{t1,t2,t3,color_2009,Boundary_2014,cheng2014global}, typically rely on color priors and handcrafted features such as positional cues and spatial structures to identify salient objects. However, due to their lack of learnability, these methods struggle to adapt to the complex and variable nature of remote sensing scenes, which limits their performance. 

Deep learning-based ORSI-SOD methods have achieved significant progress over the past decade. The introduction of convolutional neural networks (CNNs) and visual transformers (ViTs), has substantially enhanced the performance of ORSI-SOD models. CNNs, with their local receptive fields, offer high computational efficiency and strong adaptability to scale variations, making them widely adopted in ORSI-SOD tasks. For instance, SFANet\cite{SFANet_2024} enhances multiscale semantic representation through progressive flow and uncertainty-aware refinement. RAGRNet\cite{RAGNet_2024} enhances saliency detection by integrating CNN features with recurrent graph reasoning to model region–boundary interactions in complex remote sensing scenes.
Nevertheless, due to the local sliding window operations inherent in CNNs, these methods often lack the ability to model global semantic dependencies. This limitation may lead to segmentation errors and imprecise boundaries, especially when dealing with remote sensing images that contain low-contrast and large-sized objects.

To address this issue, researchers have introduced ViTs into ORSI-SOD tasks to enhance the model’s capacity for global information modeling. ViTs leverage self-attention mechanisms to perform global modeling, effectively compensating for CNNs’ limited receptive field.  
Their strong global modeling ability makes them particularly advantageous in segmentation tasks involving complex remote sensing imagery. For example, 
\textit{Gao et al.}\cite{gao_2023} introduced a novel adaptive spatial tokenization Transformer framework that produces high-quality saliency maps. Meanwhile,  
HFANet\cite{HFANet} integrates a hybrid encoder that combines CNN and Transformer to obtain features containing both global semantic relationships and local spatial details, enabling accurate detection of salient objects in complex backgrounds.
However, ViTs lack local inductive bias in modeling spatial structures\cite{ViT_lack2022,ViT_lack2024}, 
making it difficult to represent fine-grained object details and often resulting in blurred segmentation boundaries. In hybrid architectures\cite{ASNet_2024,ADSTNet_2024,HFANet}, theoretically the local detail modeling of CNNs and the global relational representation of ViTs should be complementary. However, relevant studies\cite{hunhe_lack2023,hunhe_lack2024,DPU-Former} have shown that simply combining the two fails to achieve the expected synergistic benefits. This is mainly due to significant differences in their feature representation, receptive field mechanisms, and output paradigm.

\begin{figure}[t]
  \centering
  \includegraphics[width=\columnwidth]{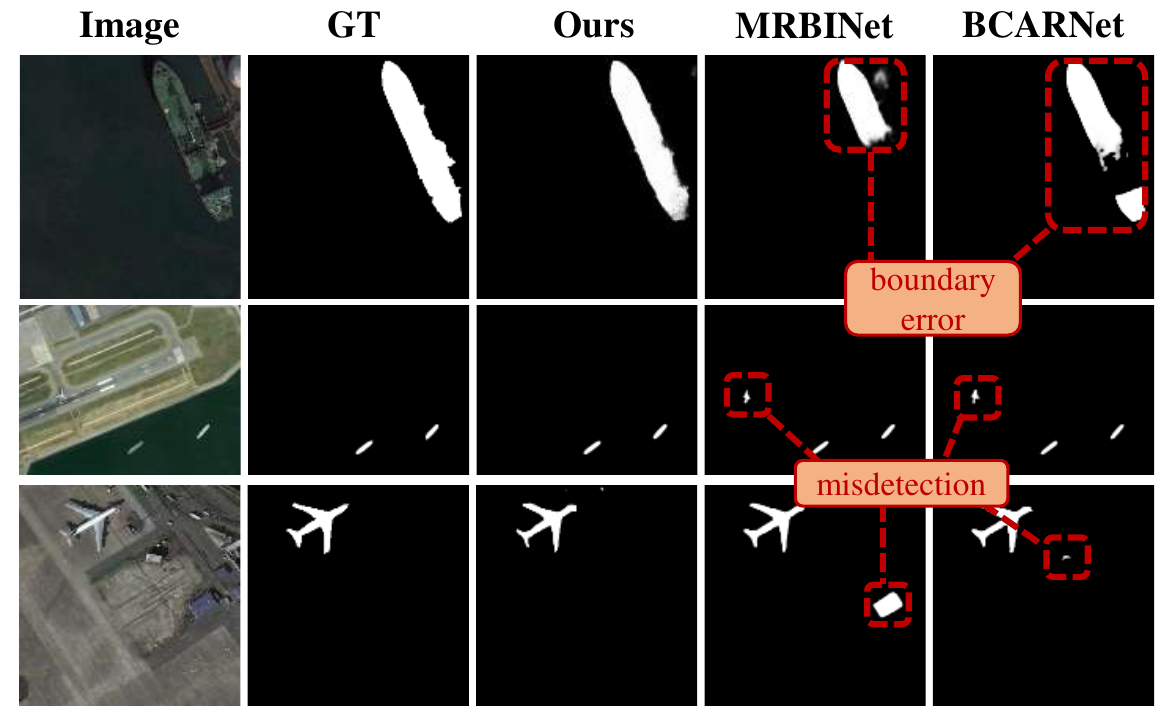}
  \caption{Our method delivers precise detection performance by balancing global and local modeling. 
  From left to right are the RGB images, ground truth (GT), results produced by our method, and those generated by representative methods, namely MRBINet\cite{MRBINet} and BCARNet\cite{BCARNet}.
  Our approach yields more accurate object localization and boundary delineation, while reducing misdetection and boundary errors.}
  \label{fig:fig2}
\end{figure}

To effectively address the aforementioned challenges, we propose a novel network architecture named adaptive state space context network (ASCNet), which is specifically designed to tackle issues such as low object-background contrast and large variations in object scales in ORSI-SOD tasks.
Inspired by recent advances in visual state space models~\cite{Mamba_2023,Vmambair_2025}, our model adopts an adaptive state space modeling strategy to enhance local context aggregation and structured feature representation within an encoder--decoder architecture.
The primary motivation lies in the fact that visual state space model employs a selective scanning algorithm to capture long-range dependencies while possessing good stability and scalability, which makes it possible to achieve an effective modeling balance between global context and local region details.
To address the challenges of blurred object boundaries and low contrast between salient objects and backgrounds in ORSIs, we propose a Multi-Level Context Module (MLCM). This module integrates multi-scale features across different representation levels and employs a graph neural network\cite{GAT_2017} to explicitly model spatial relationships and contextual dependencies among regional features. As shown in Figure \ref{fig:fig2}, by capturing cross-regional interactions and structural correlations, MLCM enables the network to better distinguish salient objects from complex backgrounds, thereby reducing segmentation errors caused by boundary ambiguity.

Meanwhile, to handle the problem of multi-scale object variations and to alleviate the limitations of sequence modeling in local region representation, we introduce an Adaptive Patchwise Visual State Space (APVSS) block. This module first feeds feature maps into a Granularity-aware Propagation Module (GPM), which enhances local structural perception by introducing global semantic conditioning and promoting consistent interpretation of regional representations.
Subsequently, a Dynamic Adaptive Granularity Scan (DAGS) operation is applied to further strengthen local modeling capability. 
Unlike conventional uniform scan strategies, DAGS partitions feature maps into resolution-aware spatial blocks and performs multi-directional scan within each block, with content-adaptive weighting to modulate the contribution of different directions for capturing local spatial dependencies. Through block-wise partitioning, direction-aware sequence construction, and integration across blocks, DAGS enhances fine-grained local feature representation while preserving global contextual awareness. This direction-adaptive modeling strategy significantly improves the network’s ability to handle scale-diverse targets and complex spatial layouts in ORSIs, ultimately leading to more accurate saliency prediction. We conduct extensive comparative experiments on two widely used benchmark datasets (i.e., ORSSD\cite{LVNet_2019} and EORSSD\cite{DFANet_2020}), and the experimental results demonstrate that ASCNet consistently outperforms existing state-of-the-art methods on ORSI-SOD tasks (as shown in Figure \ref{fig:fig1}).
Our main contributions can be summarized as follows:
\begin{enumerate}
    \item  We propose a novel network architecture named ASCNet, which enhances local region feature representation while achieving global context modeling, and introduces a topology-aware module to strengthen multi-scale interactions and boundary modeling, thereby significantly improving the detection accuracy of ORSI-SOD.
    
    \item We introduce the MLCM, which leverages a topology-aware attention mechanism to model spatial relationships and contextual dependencies across the fused multiscale features, thus enhancing the model’s structural perception capability and enabling it to effectively learn the differences between foreground and background.
    
    \item We propose APVSS, a module that integrates our proposed GPM and DAGS. It enhances the model’s capability in modeling local region details and effectively combines global context modeling with local feature learning, thereby significantly improving the accuracy and robustness of feature representation.
    
    \item Experimental results on two ORSI-SOD benchmark datasets demonstrate that our proposed method outperforms existing state-of-the-art approaches.
\end{enumerate}

\section{Related Work}
\subsection{ORSI-SOD}
Although SOD has achieved significant progress on natural scene images (NSIs), directly applying existing methods to ORSIs still faces numerous challenges. NSI-SOD methods \cite{color_2009,Boundary_2014,SODsurvey_2021,zhan2025mambasod,hesamba} are primarily designed for natural scenes from a ground-level perspective, typically relying on clear object boundaries, rich textures, and prominent semantic context. However, objects in RSIs often exhibit characteristics such as small size, low contrast, minimal inter-class variance, and strong background interference, making it difficult for existing NSI-based methods to generalize effectively in remote sensing scenarios. To address these issues, a series of SOD approaches specifically tailored for ORSIs—referred to as ORSI-SOD—have emerged in recent years, aiming to enhance detection performance in complex remote sensing environments.

Thanks to their powerful semantic representation and local feature extraction capabilities, CNNs can accurately capture fine-grained local features in ORSIs. In recent years, many CNN-based ORSI-SOD models have achieved notable progress. 
ERPNet \cite{ERPNet} employs a CNN-based encoder–decoder architecture with edge-guided position attention to progressively refine salient object localization and boundaries in optical remote sensing images.
ESGNet \cite{ESGNet} exploits edge and skeleton guidance through a two-stage CNN framework, progressively injecting structural cues into saliency features to refine object localization and preserve accurate boundaries in optical remote sensing images.
ACCoNet\cite{ACCoNet_2022} enhances saliency detection by coordinating adjacent features and capturing contextual information, leading to more comprehensive activation of salient regions. 
Similarly, Transformer-based methods have been increasingly applied to ORSI-SOD tasks, aiming to overcome the limitations of CNNs in handling complex backgrounds and weak object boundaries in optical remote sensing images. 
TLCKD-Net\cite{TLCKT-Net2024} adopts a Transformer-based encoder to effectively capture global features, addressing traditional methods’ shortcomings in handling complex backgrounds and salient objects of varying scales and shapes.
GeleNet\cite{GeleNet_2023} adopts a global-to-local paradigm based on Transformers, combining global feature extraction with local enhancement modules to more precisely locate salient objects and highlight details.
However, pure ViT architectures are limited in modeling local information, while targets in optical remote sensing images often exhibit multi-scale and complex shapes, which can easily lead to imprecise boundary segmentation. 
To address this issue, researchers have proposed hybrid architectures. 
ADSTNet\cite{ADSTNet_2024} combines the strengths of CNNs and Transformers to effectively compensate for global and local information, with boundary-guided modules further assisting in accurate detection and localization.
UDCNet\cite{UDCNet} combines CNN-based feature extraction with frequency–spatial Transformer blocks to capture complementary local and global information, while a saliency–edge joint decoder improves precise localization and boundary delineation.

By contrast, our method focuses on adaptive state space contextual modeling, which jointly captures long-range dependencies and fine-grained local structures, effectively addressing boundary ambiguity and scale variation, leading to more accurate saliency localization in complex ORSIs.

\subsection{Scan Mechanism in Visual Mamba}
Recently, the emergence of state space models (SSMs)\cite{SSM2_2021,SSM3_2022} has provided a novel paradigm for global modeling, particularly with the introduction of the Mamba\cite{Mamba_2023} model. Mamba enhances global modeling capability by introducing a selective scan mechanism and adopts a parallel scan strategy to effectively capture comprehensive contextual information across the image, facilitating better representation of long-range dependencies.
Building on this foundation, Mamba has rapidly advanced in the field of computer vision, giving rise to a variety of visual Mamba architectures\cite{VMamba_2024,Vmamba_survey_2405,Vmamba_survey2_2024}.
As a crucial component of Mamba architectures in the vision domain, efficient scan mechanisms play a key role in improving model performance and have become a focal point of current Mamba-related research. Vision Mamba employs a bidirectional scan mechanism to simultaneously process forward and backward information. VMamba\cite{VMamba_2024} introduces a cross-scan mechanism that scans images along four different paths to enable more comprehensive feature extraction, and has seen widespread adoption. 
The Hilbert scan strategy\cite{hibert2024} integrates multiple scan methods and directions, allowing it to capture local feature information from various angles. The LocalMamba \cite{localmamba} model incorporates a local scan strategy by dividing the image into multiple fixed-size, non-overlapping local windows, within which pixels are scanned independently.

\begin{figure*}[t]
  \centering
  \includegraphics[width=\linewidth]{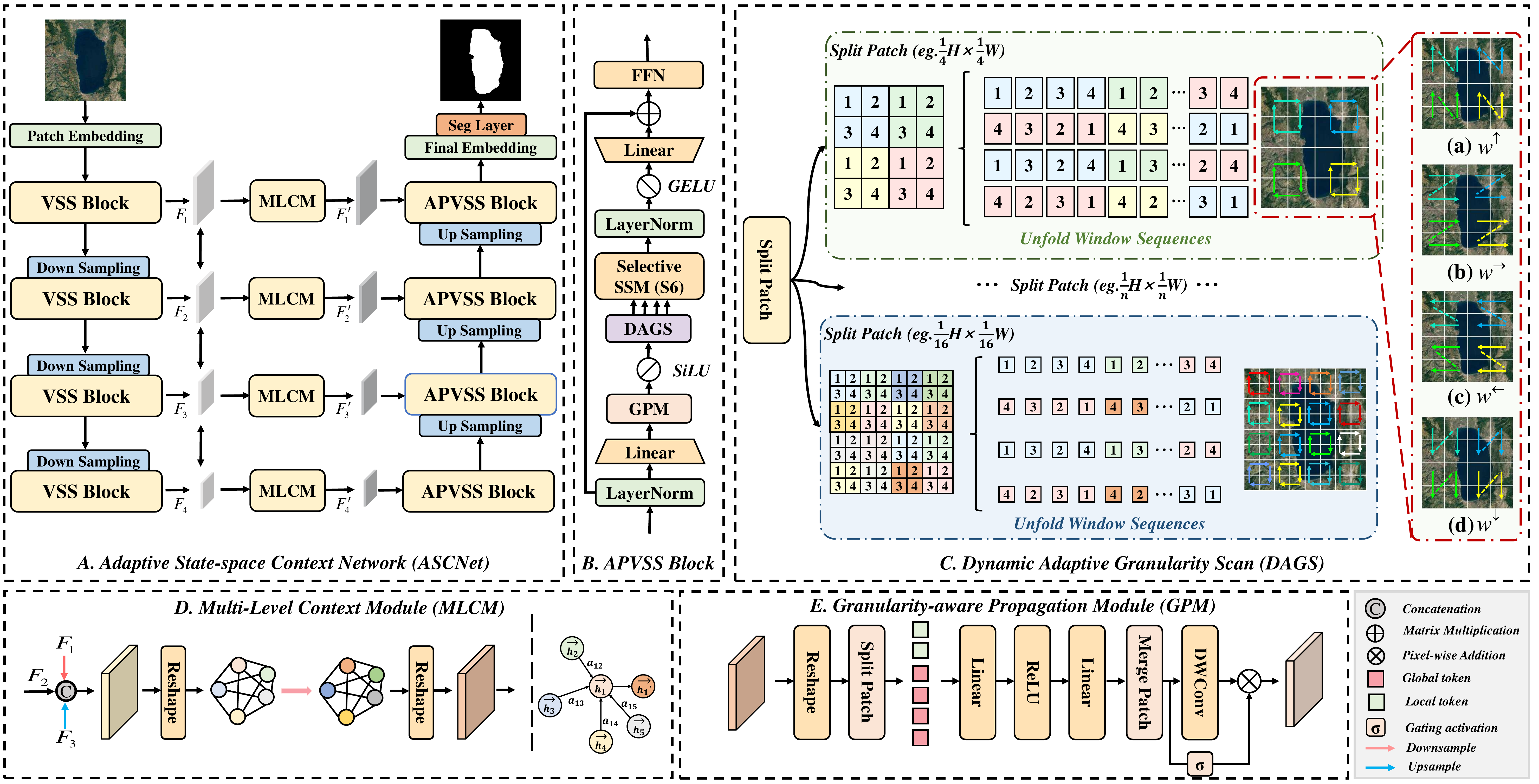}
  \vspace{-3mm}
  \caption{\textbf{Pipeline of the proposed method.} \textbf{A:} ASCNet adopts encoder-decoder architecture. The encoder is built upon the state space block. In the skip connections, a MLCM is employed to enhance the model’s representation capability through feature fusion and a topology-aware attention mechanism. Subsequently, in the decoding stage, the features processed by MLCM(D) are fed into the APVSS blocks for decoding. \textbf{B:} The proposed APVSS blocks enhance the model’s ability to capture local information through a GPM(E) and a DAGS(C), aiming for more precise boundary representation. 
  }
  \vspace{-3mm}
  \label{fig:main_framework}
\end{figure*}

Despite Mamba's impressive potential in vision applications, it still faces challenges in local modeling. In current visual Mamba models\cite{VMamba_2024,Vmambair_2025,plainmamba_2024}, a common practice is to divide 2D feature maps into patches and scan them into 1D sequences before feeding them into the SSMs for processing. However, images inherently lack a sequential structure, and this serialization conflicts with the spatial locality of image data. As a result, the dependency relationships between neighboring pixels are disrupted, leading to insufficient modeling of fine-grained structures such as edges and textures. Moreover, treating all scanning directions equally implicitly assumes isotropic spatial dependencies, which is often violated in ORSIs scenes. These limitations lead to insufficient modeling of fine-grained structures such as edges, textures, and elongated objects, ultimately hindering SSMs from generating high-quality and geometry-consistent visual representations.

\section{Methodology}
In this section, we propose a novel architecture named ASCNet, whose overall structure is illustrated in Figure \ref{fig:main_framework}. ASCNet adopts a typical encoder-decoder framework and integrates three innovative modules: MLCM, GPM, and DAGS. To clearly present our work, we first provide a summary and review of existing SSM methods in Section III-A. Then, we introduce the overall architecture of ASCNet, followed by detailed descriptions of the three core modules, including their design principles and implementation details. Finally, we explain the loss functions used in our approach.

\subsection{Preliminaries}
\textbf{State Space Model:} State space model (SSM)\cite{SSM2_2021,SSM3_2022} such as S4\cite{SSM1_2021} and Mamba\cite{Mamba_2023} are structured sequence architectures that are inspired by recurrent neural networks (RNNs) and CNNs and have the ability to scale linearly or nearly linearly with sequence length.
These models originate from continuous systems, mapping an input sequence $x(t)\in\mathbb{R}^L$ to an output sequence $y(t)\in\mathbb{R}^L$ through a hidden state $h(t)\in\mathbb{R}^N$. More specifically, SSMs are described by the following continuous-time ordinary differential equations (ODEs) in Eqs.~(\ref{eq:ssm_cont_1})--(\ref{eq:ssm_cont_2}):

\begin{align}
h'(t) &= \mathbf{A}h(t) + \mathbf{B}x(t) \label{eq:ssm_cont_1}\\
y(t) &= \mathbf{C}h(t)  \label{eq:ssm_cont_2}
\end{align}
where $h(t)$ is the current hidden state, $h'(t)$ is the updated hidden state, $x(t)$ is the current input, $y(t)$ is the output. $\mathbf{A}\in\mathbb{R}^{N \times N}$ is SSM's evolution matrix, and $\mathbf{B}\in\mathbb{R}^{N \times 1}$, $\mathbf{C}\in\mathbb{R}^{1 \times N}$ are the input and output projection matrices, respectively.

\textbf{Discrete State Space Models:}
To integrate continuous-time SSMs into deep models, they need to be discretized. This involves transforming the continuous-time functions into discrete-time sequence-to-sequence mappings. For example, this is seen in models such as S4\cite{SSM1_2021} and Mamba\cite{Mamba_2023}. One common method is zero-order hold (ZOH). It introduces a time scale parameter $\Delta$, which converts the continuous parameters $\mathbf{A}$ and $\mathbf{B}$ into their discrete versions $\overline{\mathbf{A}}$ and $\overline{\mathbf{B}}$ in Eqs.~(\ref{eq:ssm_disc_A})--(\ref{eq:ssm_disc_B}).

\begin{align}
\overline{\mathbf{A}}&=\exp(\Delta\mathbf{A}) \label{eq:ssm_disc_A}\\
\overline{\mathbf{B}}&=(\Delta\mathbf{A})^{-1}(\exp(\Delta\mathbf{A})-\mathbf{I})\cdot\Delta\mathbf{B} \label{eq:ssm_disc_B}\\
h_t&=\overline{\mathbf{A}}h_{t-1}+\overline{\mathbf{B}}x_t \label{eq:ssm_disc_state}\\
y_t&=\mathbf{C}h_t \label{eq:ssm_disc_out}
\end{align}

Based on the discretized parameters, the discrete-time state update and output equations are given by Eqs.~(\ref{eq:ssm_disc_state}) and (\ref{eq:ssm_disc_out}).

\subsection{Overall Structure}
The overall architecture of ASCNet is illustrated in Figure \ref{fig:main_framework}(A). 
ASCNet is composed of three main components: 
(1) a four-stage VSS encoder for hierarchical multi-scale feature extraction; 
(2) four Multi-Level Context Modules (MLCMs) that integrate features across different scales and model contextual relationships via topology-aware attention mechanisms; 
and (3) a four-stage APVSS decoder, which enhances local feature modeling and progressively restores spatial resolution to generate fine-grained segmentation results.
In the encoding stage, the input image 
$I \in \mathbb{R}^{H \times W \times 3}$
is first fed into a patch embedding module consisting of two convolutional layers, which transforms the image into patch-wise feature representations. 
The encoder is organized into four stages 
$s \in \{1,2,3,4\}$,
where each stage contains a $2\times$ downsampling operation followed by multiple stacked VSS blocks (the number of blocks in each stage are respectively 2, 2, 15, 2). 
Through this hierarchical design, multi-scale feature maps $F_i,\ i \in \{1,2,3,4\}$ are extracted at different spatial resolutions.
After encoding, the multi-scale features are successively processed by the MLCMs in a cascaded manner. 
Each MLCM aligns and fuses features from different levels and employs a topology-aware attention mechanism to explicitly model long-range dependencies among regions in the fused feature maps. 
As a result, the enhanced feature representations 
$F_i'$ are obtained, which effectively capture global contextual relationships. 
This design enables the network to reliably highlight salient objects under complex background interference and significantly improves the discriminative capability for salient regions.
In the decoding stage, ASCNet consists of four decoding levels. Each decoding stage contains two stacked Adaptive
Patchwise Visual State Spac (APVSS) blocks to further strengthen local feature modeling within each scale. 
The deepest decoding stage takes the high-level encoded features as input, while the remaining stages integrate the context-enhanced features at the corresponding scale with the upsampled intermediate representations from the previous decoding stage along the channel dimension. 
Through progressive feature fusion and refinement, the decoder gradually recovers spatial details and produces high-resolution segmentation predictions.
\subsection{Multi-Level Context Module}
In ORSI-SOD, representations from a single abstraction level are often insufficient for reliable saliency discrimination in complex and low-contrast scenes. High-level features provide coarse semantic cues, while low-level features preserve fine-grained spatial structures and boundary details, making multi-level contextual aggregation essential for accurate salient object localization.
Moreover, although optical ORSIs often exhibit visually cluttered backgrounds, salient objects tend to present spatially clustered distributions with consistent visual patterns across regions and representation levels. This property suggests that saliency discrimination can benefit from explicitly modeling relationships among spatially separated yet semantically related regions across multiple levels. Motivated by this observation, we introduce the Multi-Level Context Module (MLCM), which aims to aggregate contextual cues across different representation levels and capture region-level dependencies through structured relational modeling.

As illustrated in Figure \ref{fig:main_framework}(D), MLCM operates on a set of feature representations  $F_i$, where $i\in \{1,2,3,4\}$ extracted from different encoding stages, each corresponding to a distinct abstraction level. For each level, a multi-scale feature fusion operation is first employed to align and integrate information from its adjacent representation levels into a unified representation at the target resolution. This design enables each level to access complementary contextual cues from other abstraction levels. The fusion process is formulated as:
\begin{equation}
F_i^{c}=Conv_{1\times 1}(Concat(\phi_1(F_{i-1}),\phi_3(F_i),\phi_4(F_{i+1}))
\end{equation}
where $\phi_i(\cdot)$ denotes resolution alignment via upsampling or downsampling, $Concat$ represents channel-wise concatenation, and $Conv_{1\times 1}$ compresses the fused representation into a compact feature space.

Meanwhile, to reduce the number of parameters processed by the subsequent topology-aware attention networks operations, the module downsamples the fused feature, the feature $F^{c}_i\in \mathbb{R}^{h\times w\times c}$ is fed into a convolutional layer with a kernel size of $s\times s$ and a stride of $s$ ($s=3$).
Then, each spatial location in the downsampled feature map $F^{c'}_i\in \mathbb{R}^{h/s\times w/s\times c}$ (corresponding to an $s\times s$ local region in the original feature) is treated as a topological node, and the channel vector at that location is used as the node feature $\vec{f}_k\in \mathbb{R}^d$.
Therefore, the downsampled feature map can be rearranged into a region-level feature representation with explicit spatial topological structure: $F_{in}=\{\vec{f}_1,\vec{f}_2,...,\vec{f}_n \},n=h/s\times w/s$.
To transform the input features into higher-level representations and enhance the expressiveness of the network, we apply a shared learnable weight matrix $W\in \mathbb{R}^{d'\times d}$ to linearly transform each node. Then, a shared self-attention weight matrix $l\in \mathbb{R}^{2d'\times 1}$ is used to compute the attention coefficients between node pairs.
The attention coefficient $a_{ij}$ is computed only for nodes $j \in \mathcal{N}_i$, where $\mathcal{N}_i$ denotes the first-order neighborhood of node $i$ in the reorganized region-level representation. 
Specifically, $\mathcal{N}_i$ consists of the spatially adjacent nodes in the 2D grid of the downsampled feature map, implemented using a fixed local neighborhood. 
This localized attention design enables efficient relational modeling while preserving local spatial consistency.
To better compare the attention coefficients $a$ between different nodes, we apply the softmax function to normalize all the selected $j$. Therefore, the formula for $a_{ij}$ is:

\begin{equation}
a_{ij}=\frac{exp(\sigma(l^T[W\vec{f}_i||W\vec{f}_j]))}{\sum_{n\in\mathcal{N}_i}exp(\sigma(l^T[W\vec{f}_i||W\vec{f}_n]))}
\end{equation}
Here, $T$ denotes the transpose operation, $||$ represents the concatenation operation, and $\sigma$ is the LeakyReLU activation function. Assuming $\delta$ is the ELU activation function, the final output feature can be expressed as:
\begin{equation}
\vec{f}'_{i}=\delta(\sum_{n\in\mathcal{N}_i}a_{ij}W\vec{f}_j)
\end{equation}
Finally, the feature maps enhanced by two consecutive topological dependency modeling steps are reshaped back to the spatial domain and projected to the original resolution through a deconvolution followed by an upsampling operation, denoted as 
$F_{out}=\{\vec{f}'_1,\vec{f}'_2,...,\vec{f}'_n\},\vec{f}'_n\in\mathbb{R}^d$, yielding the final output of the MLCM.
This output feature effectively integrates multi-level contextual information and region-level relational dependencies, serving as a discriminative representation for subsequent saliency decoding.

\subsection{Adaptive Patchwise Visual State Space Block}
We propose the Adaptive Patchwise Visual State Space (APVSS) module, which is designed on top of the original VMamba module to enhance the representation of local region information and significantly improve the model's ability to capture local region dependencies. The structure of APVSS is illustrated in Figure \ref{fig:main_framework} (B) and consists of two main components: GPM and DAGS.
\subsubsection{Granularity-aware Propagation Module}
In optical remote sensing salient object detection, accurate saliency discrimination requires regional representations to be interpreted under globally consistent semantic cues. To this end, we propose the Granularity-aware Propagation Module (GPM), which performs semantic conditioning on regional representations prior to sequence construction, thereby guiding subsequent state space modeling with stable global context (as illustrated in Figure \ref{fig:main_framework}(D)).
The GPM is embedded into the adaptive state space decoding pipeline and applied before sequence construction. Its role is to establish a semantically conditioned perceptual space in which regional representations can be interpreted more consistently across complex scenes.

Given an input feature map $F_i \in \mathbb{R}^{h \times w \times c}$, GPM first partitions it into a set of non-overlapping local patches. Each patch is flattened into a sequence of tokens $T_i \in \mathbb{R}^{p^2 \times C}$, where $p$ ($p=16$) denotes the patch size and $i$ indexes the patch.
To inject global semantic priors into regional perception, we introduce 2 learnable global tokens $G \in \mathbb{R}^{2 \times C}$ and concatenate them with the patch tokens. 
The concatenated tokens are then processed by a global semantic interaction operator, which enables feature mixing and information propagation under global semantic conditioning:
\begin{equation}
\hat{T}_i = [G; T_i] \in \mathbb{R}^{(2 + p^2) \times C}
\end{equation}
\begin{equation}
Z_i = (LN(\hat{T}_i)W_1)\odot\sigma(Pool_g(LN(\hat{T}_i))W_2)
\end{equation}
where $W_1 ,W_2\in\mathbb{R}^{C\times C}$are learnable projection matrices, $Pool_g(\cdot)$denotes global pooling over the token dimension (including global tokens), and $\sigma(\cdot)$is the sigmoid activation function.
Through this semantic interaction process, the global tokens serve as semantic anchors that aggregate image-level contextual information and generate global conditioning signals to regulate regional feature responses. As a result, patch tokens are interpreted under globally consistent semantic cues, which enhances the separability between salient objects and background regions in complex spatial layouts. After semantic interaction, only the updated patch tokens are retained:
\begin{equation}
T'_i = Z_i[2 + 1 : 2 + p^2]
\end{equation}

The refined patch tokens are subsequently reshaped back into two-dimensional feature blocks and reassembled to form an enhanced feature map $X'$. On top of $X'$, a gated convolutional feed-forward network is applied to further regulate feature responses. Specifically, depthwise separable convolution\cite{DWConv2017} captures local structural patterns, while a gating mechanism adaptively modulates feature activation strength:
\begin{equation}
Y = DWConv(X') \odot \sigma(X')
\end{equation}

The resulting representation integrates region-level structural information with global semantic conditioning and is forwarded to subsequent sequence construction and state space modeling stages.

\subsubsection{Dynamic Adaptive Granularity Scan
}
Compared with natural images, salient targets in ORSIs are often embedded in cluttered backgrounds and exhibit diverse geometric patterns across scales and regions. Consequently, effective state space modeling for salient object detection in ORSIs requires the sequence construction process to adapt not only to the heterogeneous spatial organization of visual representations across different resolutions, but also to the diverse geometric patterns within local regions.
Feature maps at different scales exhibit distinct levels of geometric complexity: lower-resolution representations mainly encode coarse semantic layouts with compact spatial structures, while higher-resolution ones preserve richer geometric details and fine-grained spatial variations \cite{ECCV_2018,CVPR_2017}. Moreover, even within the same resolution, local regions may present different directional structures and spatial arrangements. However, applying a uniform scanning strategy across all resolutions and spatial locations implicitly assumes homogeneous spatial organization, which is inconsistent with the resolution- and content-dependent nature of dense visual representations. Motivated by this observation, we propose Dynamic Adaptive Granularity Scan (DAGS), a resolution-aware and content-adaptive scanning strategy embedded within the adaptive state space decoder block to better align sequence construction with multi-resolution and region-specific geometric characteristics.

DAGS addresses this issue by dynamically adjusting both the spatial granularity and the relative importance of directional scans during sequence construction. Given an input feature map $I \in \mathbb{R}^{h \times w \times c}$, DAGS first partitions the representation into spatial blocks according to its resolution, enabling resolution-aware modeling of spatial dependencies within the decoder block. For low-resolution feature maps (e.g., $\frac{1}{16}h \times \frac{1}{16}w$), where spatial structures are relatively compact and global semantic layouts dominate, DAGS applies four-directional global scanning to preserve long-range contextual interactions. In contrast, for medium- and high-resolution feature maps (e.g., $\frac{1}{8}h \times \frac{1}{8}w$, $\frac{1}{4}h \times \frac{1}{4}w$, and $\frac{1}{2}h \times \frac{1}{2}w$), the feature maps are divided into multiple non-overlapping sub-regions (e.g., 4, 16, and 64 regions, respectively), allowing localized modeling of fine-grained spatial structures.
For each sub-region $F_i$, DAGS performs four directional scan operations in parallel:
\begin{equation}
\begin{split}
Scan_4(F_i) = [&\, Scan^{\rightarrow}(F_i),\, Scan^{\downarrow}(F_i), \\
              &\, Scan^{\leftarrow}(F_i),\, Scan^{\uparrow}(F_i)]
\end{split}
\end{equation}
Instead of treating all directional scans equally, DAGS introduces a lightweight content-adaptive weighting mechanism to modulate the contribution of each directional sequence. Specifically, a direction-aware response vector $g_i\in \mathbb{R}^4$ is predicted from the aggregated sub-region feature using global average pooling followed by a learnable projection. The directional importance weights are then obtained via Softmax normalization:

\begin{equation}
\mathbf{g}_i = \phi\!\left(GAP(F_i)\right),
\end{equation}

\begin{equation}
w_i^{d} =
\frac{\exp\left(g_i^{d}\right)}
{\sum_{d' \in \{\rightarrow,\,\downarrow,\,\leftarrow,\,\uparrow\}}
\exp\left(g_i^{d'}\right)},
\quad d \in \{\rightarrow,\,\downarrow,\,\leftarrow,\,\uparrow\}
\end{equation}
where $w_i^{d}\in \mathbb{R}^4$, $\phi(\cdot)$ denotes a learnable projection, and $GAP(\cdot)$ is global average pooling. Each directional scan sequence is then reweighted accordingly:

\begin{equation}
\tilde{S}_i^{d} = w_i^{d} \cdot Scan^{d}(F_i),
\quad d \in \{\rightarrow,\,\downarrow,\,\leftarrow,\,\uparrow\}
\end{equation}

Finally, the reweighted directional scan sequences from all sub-regions are concatenated in a fixed spatial and directional order to form the complete sequence representation fed into the state space model. By preserving the four-directional scanning paradigm while enabling content-adaptive modulation at the sub-region level, DAGS allows the state space model to flexibly emphasize geometrically consistent directions across different resolutions and spatial locations, effectively balancing global contextual modeling and fine-grained directional sensitivity without introducing additional architectural stages.

\subsection{Loss Function}
We define an integrated loss function to constrain the entire training process. Our loss function consists of two components: binary cross-entropy (BCE) loss and intersection over union (IoU) loss.
Specifically, we calculate the losses for saliency predictions at four different scales: $P_1$, $P_2$, $P_3$, and $P_4$, where $P_1$ is the final saliency map while $P_2$, $P_3$, and $P_4$ are intermediate-scale saliency maps are used for auxiliary training. The total loss function is defined as:
\begin{equation}
\mathcal{L}_{total}=\sum\limits_{i=1}^{4}{
\lambda_i(\mathcal{L}_{BCE}(P_i,G_i)+\mathcal{L}_{IoU}(P_i,G_i))}
\end{equation}
where $\lambda_i$ are the weight coefficients, $P_i$ represents the saliency map, and $G_i$ is the corresponding ground truth map. In this paper, we set $\lambda_1=\lambda_2=\lambda_3=\lambda_4$ to ensure that the feature maps at all scales contribute equally to the final loss function.

BCE loss is used to ensure pixel-wise classification accuracy. For each predicted saliency map $P_i$ and its corresponding ground truth map $G_i$, the binary cross-entropy loss is defined as:
\begin{equation}
\mathcal{L}_{BCE}(P_i,G_i)=-\frac{1}{N}\sum\limits_{j=1}^N{[G^j_ilog(P^j_i)
+(1-G^j_i)log(1-P^j_i)]}
\end{equation}
where $N$ represents the total number of pixels, and $P^j_i$ and $G^j_i$ represent the values at the $j$-th pixel of the predicted map and the ground truth map, respectively.
IoU loss is used to maximize the overlap between the predicted region and the ground truth region. The IoU loss is defined as:
\begin{equation}
\mathcal{L}_{IoU}(P_i,G_i)=1-\frac{\sum{(P_i\odot G_i)+\epsilon}}
{\sum{(P_i+G_i-P_i\odot G_i)+\epsilon}}
\end{equation}
where $\odot$ denotes element-wise multiplication, and $\epsilon$ is a smoothing term used to prevent division by zero errors.

\section{Experiments}
\subsection{Experimental Setup}
\textbf{Datasets.}
To evaluate the effectiveness of the method, we trained and tested it on two public benchmark datasets for ORSI-SOD: ORSSD\cite{LVNet_2019} and EORSSD\cite{DFANet_2020}. 

The ORSSD dataset is the first publicly released dataset in the field of ORSI-SOD. This dataset contains a total of 800 high-quality visible light remote sensing images, which include eight types of remote sensing objects, such as islands, ships, and vehicles. Of these, 600 images are allocated to training sets for algorithm training and optimization, while the remaining 200 images are designated as test sets to evaluate the model's performance and generalization ability.

The EORSSD dataset is an enhanced version of the ORSSD dataset, featuring significant improvements in both data volume and diversity. Compared to the original ORSSD, the data volume of EORSSD has more than doubled, offering a richer array of training and testing samples. The dataset comprises 2,000 high-resolution images along with corresponding pixel-level ground truth, encompassing both high-resolution satellite images and low-resolution aerial images. It includes nine distinct remote sensing objects: ships, airplanes, cars, buildings, water, islands, roads, none, and others. Of these, the training set contains 1,400 images, while the test set consists of 600 images.

\begin{table*}[htbp]
\centering
\caption{Quantitative comparison results in terms of $F_{\beta}^{max} $, 
$E_{\xi}^{max}$, 
$S_{\alpha}$ and $MAE$ score are presented on two benchmark datasets. In the tables, \( \uparrow \) and \( \downarrow \) indicate that higher and lower values are preferred, respectively. The best result in each column is highlighted in bold, and the second-best result is underlined.
}
\renewcommand{\arraystretch}{1.2}
\begin{tabular}{c|c|cccc|cccc}
\hline\hline
\multirow{2}{*}{\centering Method} & \multirow{2}{*}{\centering Year} &
\multicolumn{4}{c|}{EORSSD} & \multicolumn{4}{c}{ORSSD}\\
\cline{3-10}
&&$F_{\beta}^{max} \uparrow$ & $E_{\xi}^{max} \uparrow$ & $S_{\alpha} \uparrow$ & $MAE\downarrow$ 
&$F_{\beta}^{max} \uparrow$ & $E_{\xi}^{max} \uparrow$ & $S_{\alpha} \uparrow$ & $MAE \downarrow$ 
\\
\hline
EMFINet & 2021 & .8721 &  .9712  & .9292 & .0084 & .9002  & .9737 & .9366 & .0109 \\
ERPNet&2022&.8744&.9665&.9254&.0082&.9036&.9738&.9352&.0114\\
ESGNet&2023&.8907&.9665&.9374&\underline{.0051}&.9220&.9858&.9484&.0065\\
BSCGNet&2023&.8809&.9693&.9296&.0079&.9101&.9784&.9411&.0093\\
ACCoNet&2023&.8822&.9759&.9303&.0067&.9149&.9819&.9428&.0087\\
SFANet&2024&.8834&.9770&.9351&.0058&.9192&.9830&.9453&.0077\\
RAGRNet&2024&.8853&.9785&.9364&.0057&.9242&.9861&.9507&.0066\\
UDCNet&2024&.8821&.9775&.9311&.0056&.9239&.9850&.9497&.0068\\
BCARNet&2025&.8871&.9762&.9361&.0054&.9196&.9833&.9465&.0071\\
MRBINet&2025&.8854&.9766&.9355&.0056&.9199&.9851&.9474&.0069\\
THMNet&2025&\underline{.8981}&\underline{.9827}&\underline{.9432}&.0054&.9251&.9843&\underline{.9503}&.0071\\
LGIPNet&2025&.8837&.9761&.9339&.0065&.9149&.9806&.9450&.0082\\
DPU-Former&2025&.8931&.9816&.9402&.0056&\underline{.9263}&\underline{.9868}&.9412&\underline{.0062}\\
\hline
\textbf{ASCNet (ours)}&---&\textbf{.9009}&\textbf{.9838}&\textbf{.9449}&\textbf{.0048}&\textbf{.9383}&\textbf{.9924}&\textbf{.9582}&\textbf{.0058}\\
\hline\hline
\end{tabular}

\label{tab:exper}
\end{table*}

\begin{figure*}[!t]
  \centering
  \includegraphics[width=\linewidth]{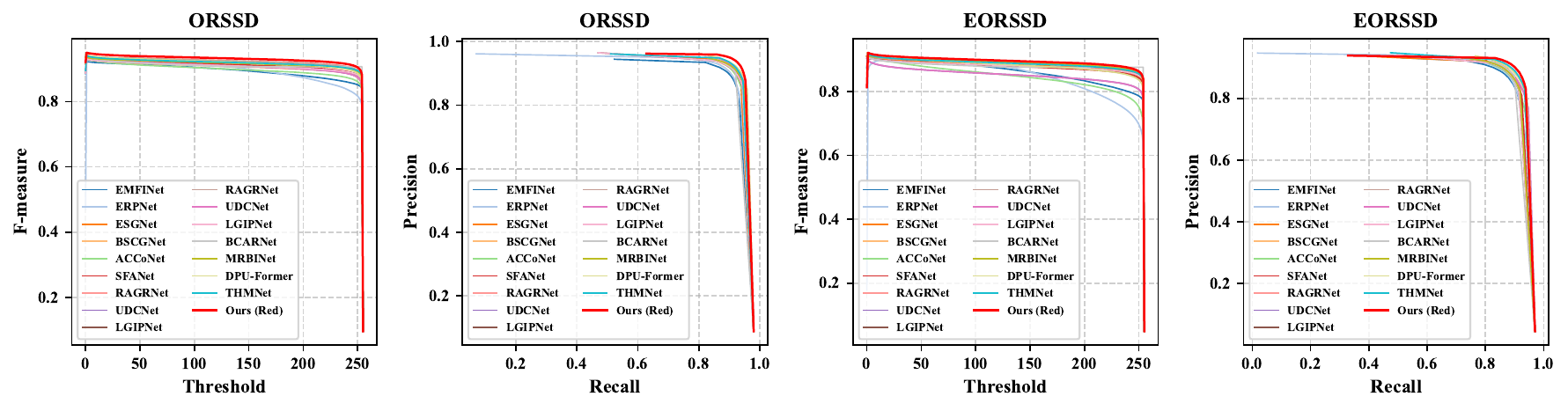}
  \caption{The quantitative comparisons between our method and other ORSI-SOD approaches on the ORSSD\cite{LVNet_2019} and EORSSD\cite{DFANet_2020} datasets are presented. }
  \label{fig:pr_Fmeasure}
\end{figure*}

\textbf{Implementation Details.} In this study, we implemented the proposed method using the PyTorch framework and conducted all experiments on an RTX 4090. We set the learning rate to 1e-4, the batch size to 4, and employed the AdamW optimization strategy for training. To ensure consistency in the input data, each image was resized to 384×384 pixels before being fed into the model. Additionally, to effectively mitigate overfitting during training, we introduced a series of data augmentation techniques, including random scaling and cropping, random image transformations, and normalization. For the encoder, we adopted VMamba\cite{VMamba_2024} initialized with publicly available pre-trained weights and fine-tuned it end-to-end on the target datasets. Finally, we trained the model for 100 epochs on the ORSSD and EORSSD datasets.

\begin{figure*}[!t]
  \centering
  \includegraphics[width=\linewidth]{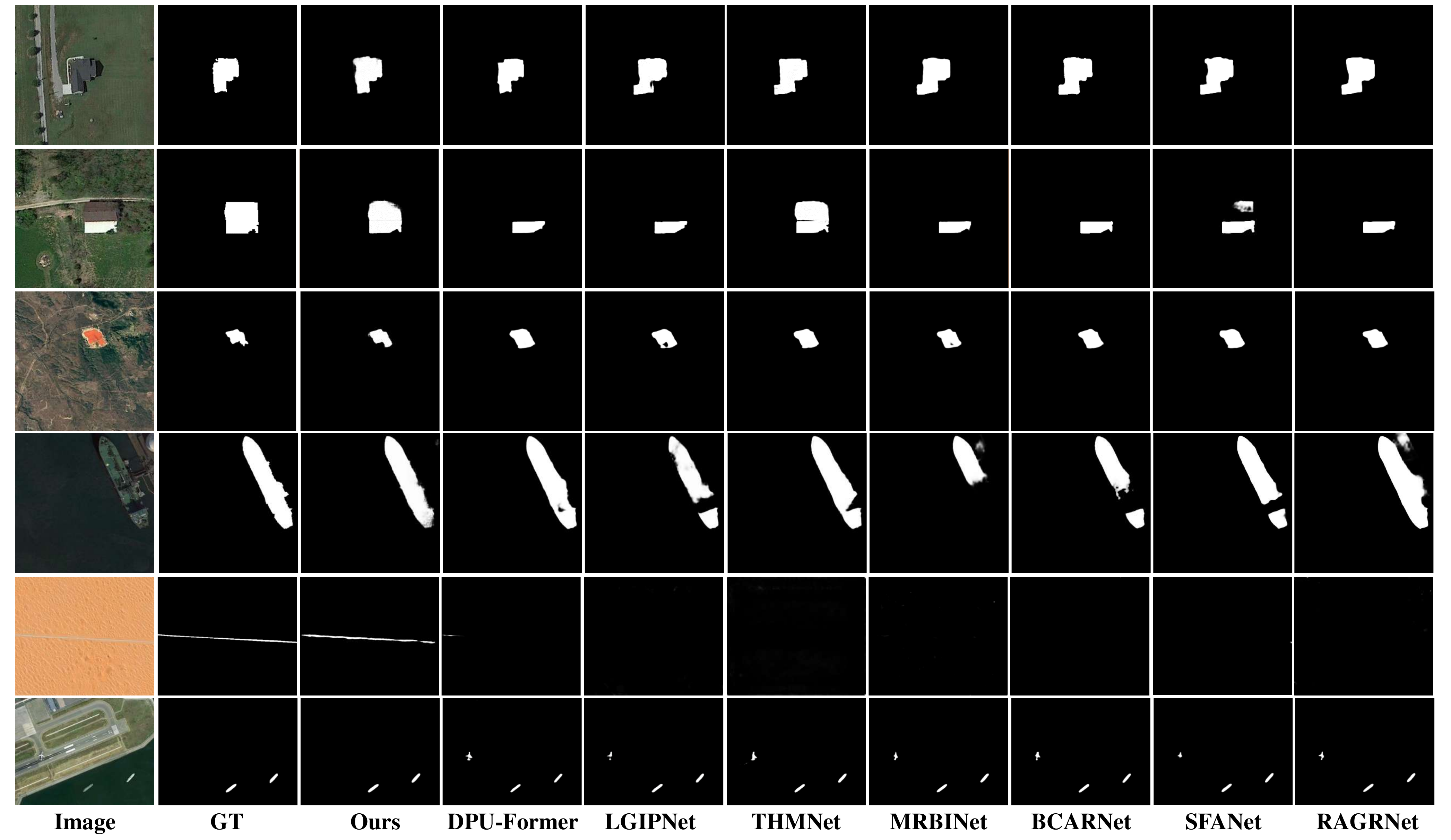}
  \caption{The qualitative comparison between our method and other seven SOTA approaches across various challenging scenarios. Our method yields more accurate object localization and clearer boundaries across different scenarios. See the zoomed-in view for better visualization.}
  \label{fig:Visualization}
\end{figure*}
\begin{figure*}[t]
  \centering

  \begin{minipage}{0.48\textwidth}
    \centering
    \includegraphics[width=\linewidth]{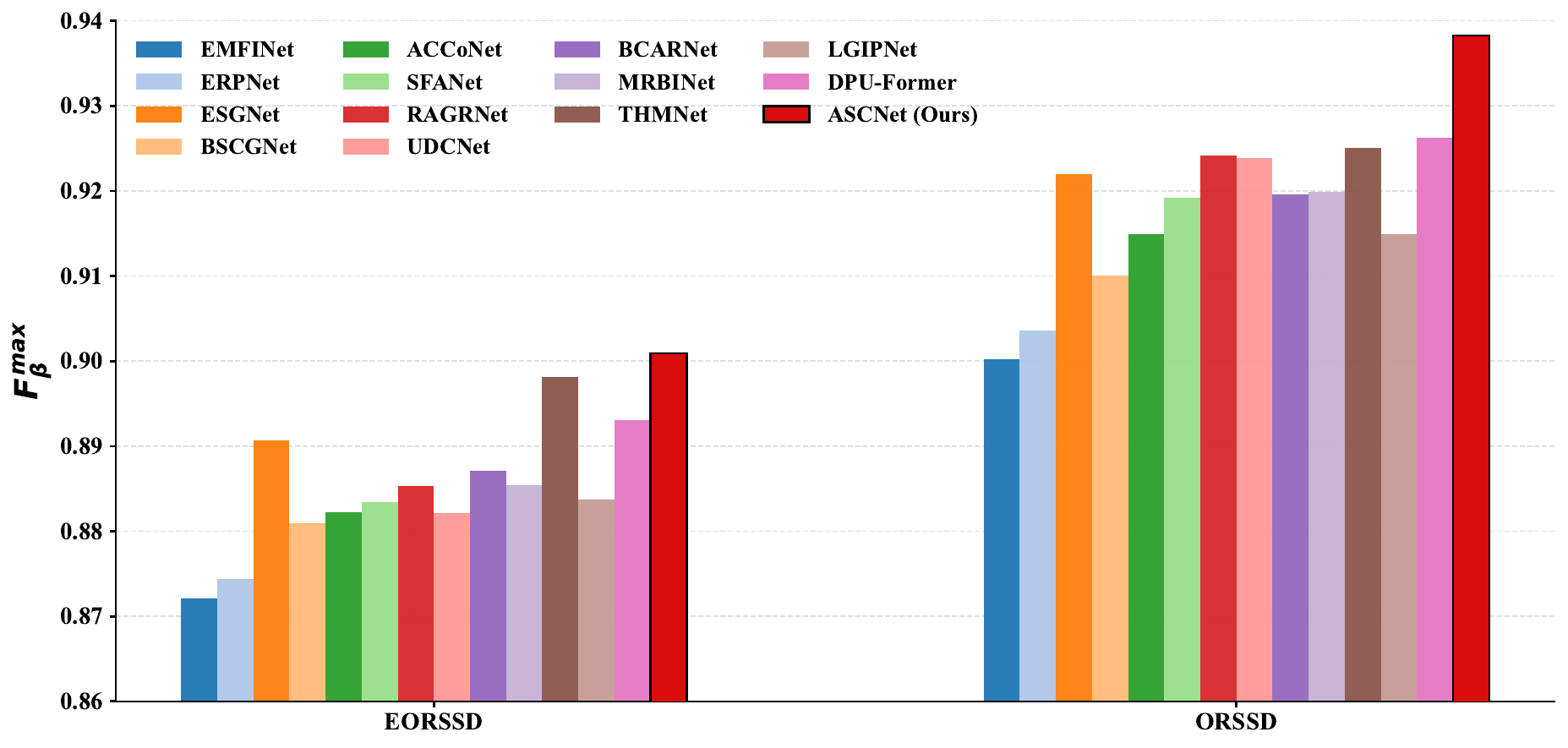}
    \vspace{-0.2em}
    {\small (a)}
  \end{minipage}
  \hfill
  \begin{minipage}{0.48\textwidth}
    \centering
    \includegraphics[width=\linewidth]{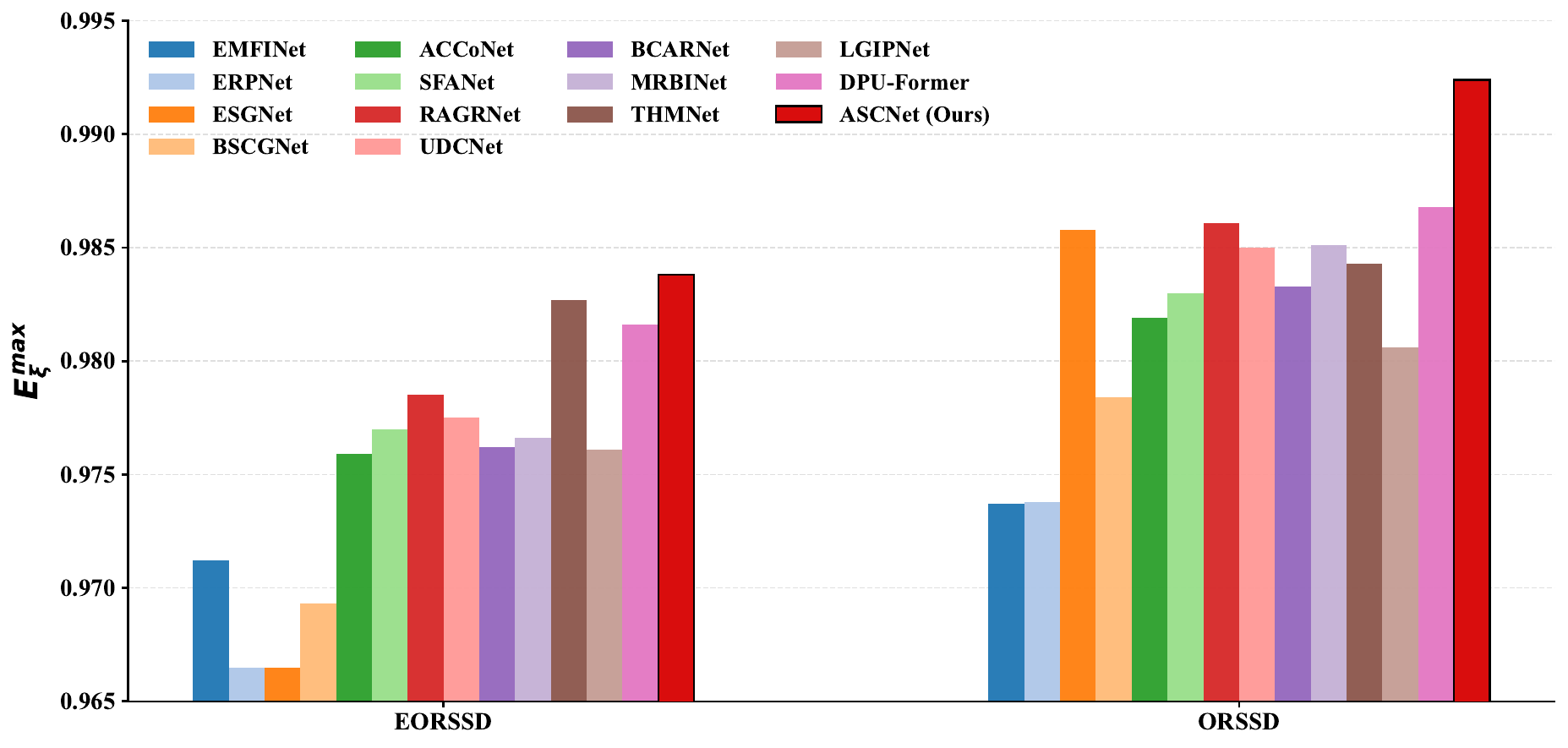}
    \vspace{-0.2em}
    {\small (b)}
  \end{minipage}

  \vspace{0.8em}

  \begin{minipage}{0.48\textwidth}
    \centering
    \includegraphics[width=\linewidth]{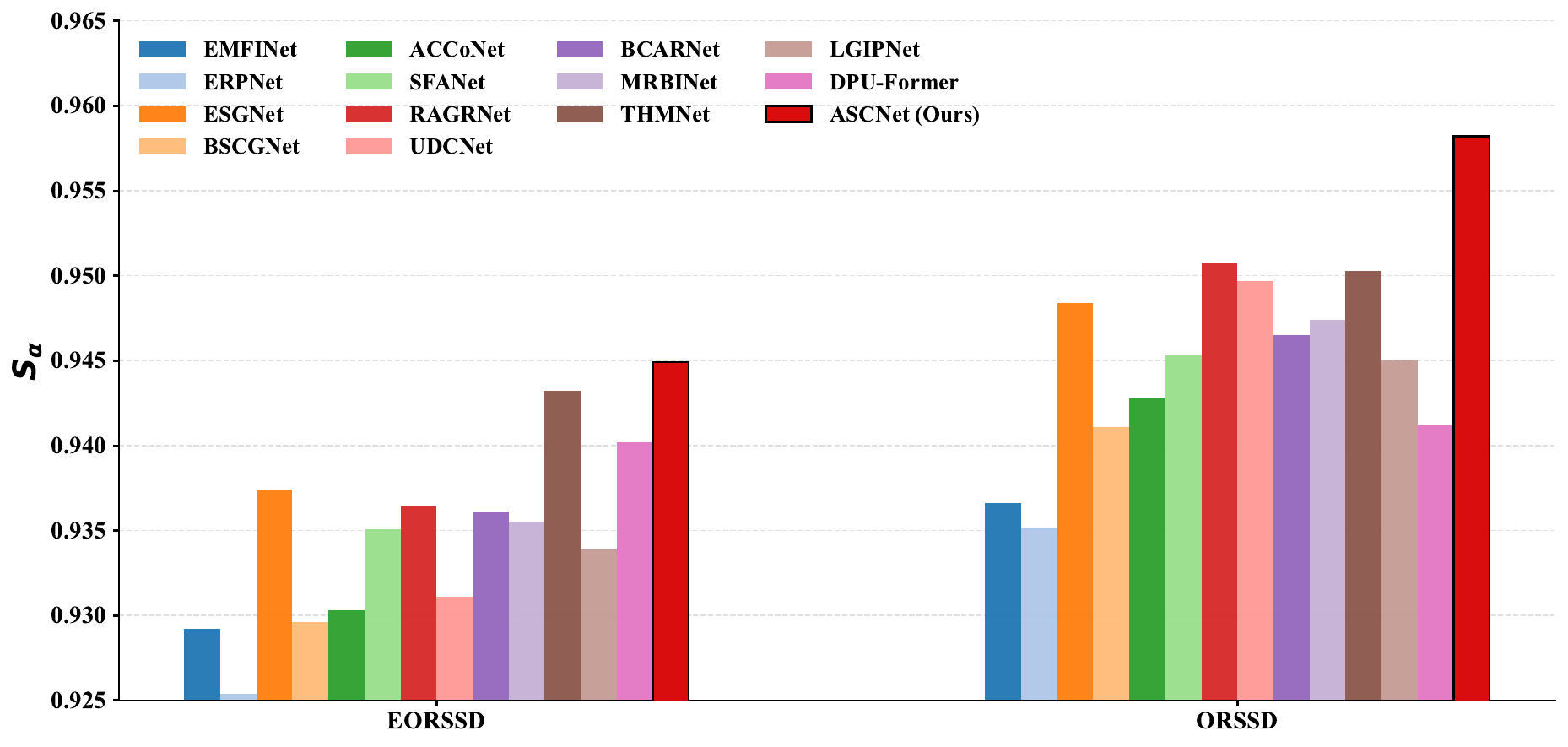}
    \vspace{-0.2em}
    {\small (c)}
  \end{minipage}
  \hfill
  \begin{minipage}{0.48\textwidth}
    \centering
    \includegraphics[width=\linewidth]{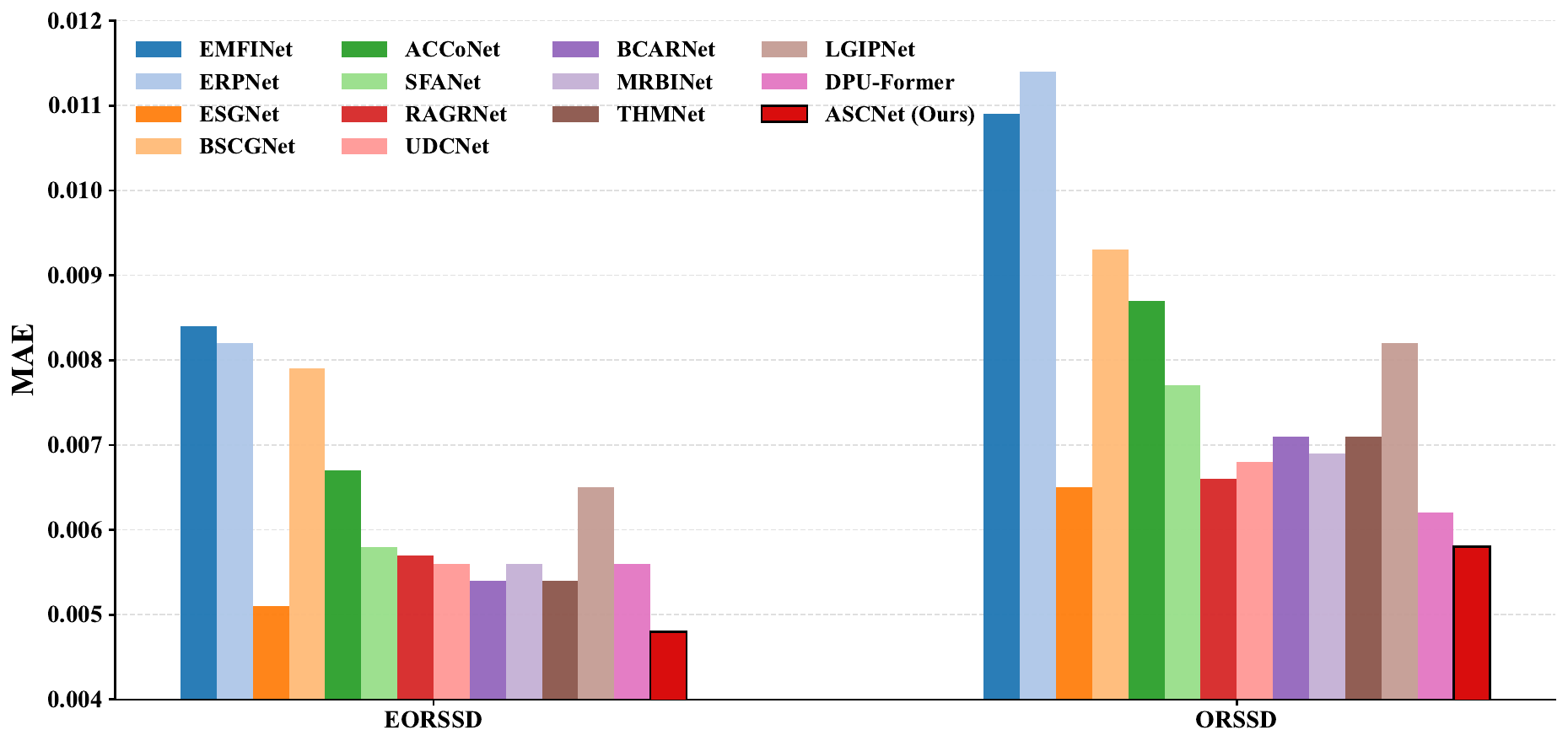}
    \vspace{-0.2em}
    {\small (d)}
  \end{minipage}

  \caption{Quantitative comparison on the EORSSD and ORSSD datasets.
  (a) $F_{\beta}^{\max}$,
  (b) $E_{\xi}^{\max}$,
  (c) $S_{\alpha}$,
  and (d) MAE.}
  \label{fig:quant}
\end{figure*}

\textbf{Evaluation Metrics.}
To comprehensively evaluate the performance of the proposed method, we adopt four widely used evaluation metrics to assess the performance of the model.
These metrics include the Mean Absolute Error ($MAE$), S-measure ($S_{\alpha}$) 
and maximum F-measure ($F^{max}_{\beta}$). Specifically, $S_{\alpha}$ quantifies the structural similarity between the predicted saliency map and the ground truth. $F_{\beta}$ is  the weighted harmonic mean of precision and recall, where $\beta^{2}$ is set to 0.3 to place greater emphasis on precision over recall.
The $MAE$ calculates the average pixel-wise error between the prediction and the ground truth, offering a direct measure of prediction deviation. 
Furthermore, $E^{max}_{\xi}$ evaluate the E-measure using maximum values. 
Similarly, $F^{max}_{\beta}$ assess the F-measure using maximum values, emphasizing the trade-off between precision and recall.

\subsection{Performance Comparisons}
 We compare our method with 13 state-of-the-art methods specifically designed for the ORSI-SOD task, including EMFINet\cite{EMFINet}, ERPNet\cite{ERPNet}, ESGNet\cite{ESGNet}, BSCGNet\cite{BSCGNet}, ACCoNet\cite{ACCoNet_2022}, SFANet\cite{SFANet_2024}, RAGRNet\cite{RAGNet_2024}, UDCNet\cite{UDCNet}, BCARNet\cite{BCARNet}, MRBINet\cite{MRBINet}, THMNet\cite{THMNet}, LGIPNet\cite{LGIPNet}, and DPU-Former\cite{DPU-Former}. To ensure a fair evaluation, we adopt the saliency maps generated using the official implementations or pretrained models provided by the original authors, and all methods are evaluated using the same evaluation toolkit and protocols on the ORSSD and EORSSD datasets

\textbf{Quantitative Comparisons.}
Figure~\ref{fig:quant} presents a visual quantitative comparison on the EORSSD and ORSSD datasets in terms of (a) $F_{\beta}^{\max}$, (b) $E_{\xi}^{\max}$, (c) $S_{\alpha}$, and (d) MAE. As shown in the figure, our method consistently achieves superior performance across all evaluation metrics on both datasets.
The detailed quantitative results of our proposed ASCNet and other state-of-the-art methods are summarized in Tab.~\ref{tab:exper}. It is evident that our model achieves SOTA performances.
On the EORSSD dataset, the quantitative comparison results of our method across four evaluation metrics are presented at the bottom of the table. It can be observed that our approach achieves the best performance on all key indicators, including $F_{\beta}^{max}$, $E_{\xi}^{max}$, $S_{\alpha}$, and $MAE$. Specifically, compared with the second-best method, our approach achieves performance gains of 
0.28\% on $F_{\beta}^{max}$, 
0.11\% on $E_{\xi}^{max}$, 
and 0.17\% on $S_{\alpha}$, further highlighting its superiority in salient object detection tasks.
On the ORSSD dataset, our method also achieves superior performance. Specifically, it also achieved the best results across all metrics, 
compared to the second-best method, our approach achieves a 1.2\% improvement on metric $F_{\beta}^{max}$, a 0.56\% improvement on metric $E_{\xi}^{max}$, and a 0.79\% improvement on metric $S_{\alpha}$, 
further validating the effectiveness and robustness of our approach. 

Figure \ref{fig:pr_Fmeasure} illustrates the PR and F-measure curves on the ORSSD and EORSSD datasets. Higher curves indicate better performance. Compared with other methods, our approach consistently achieves higher precision and recall. Specifically, our PR curve is closer to the top-right corner, demonstrating superior performance across thresholds. Similarly, our F-measure curve outperforms all competing methods. These results collectively highlight the effectiveness and robustness of our method under various conditions.

\begin{table*}[!h]
\centering
\caption{Comparison of different module ablation settings on ORSSD and EORSSD datasets.
The best result in each column is highlighted in bold.}
\label{tab:module_ablation}
\resizebox{0.8\textwidth}{!}{
\begin{tabular}{c|ccc|cccc|cccc}
\toprule
\multirow{2}{*}{NO.} &
\multicolumn{3}{c|}{Method} &
\multicolumn{4}{c|}{ORSSD} &
\multicolumn{4}{c}{EORSSD} \\
\cmidrule(lr){2-4}
\cmidrule(lr){5-8}
\cmidrule(lr){9-12}
& MLCM & GPM & DAGS
& $S_{\alpha}\!\uparrow$ & $F_{\beta}^{max}\!\uparrow$ & $E_{\xi}^{max}\!\uparrow$ & $MAE\!\downarrow$
& $S_{\alpha}\!\uparrow$ & $F_{\beta}^{max}\!\uparrow$ & $E_{\xi}^{max}\!\uparrow$ & $MAE\!\downarrow$ \\
\midrule
(a) &  &  &  &
.9049 & .879 & .9663 & .0178 &
.9078 & .8417 & .9566 & .0100 \\

(b) & \cmark &  &  &
.9284 & .8981 & .9761 & .0121 &
.9249 & .8681 & .9698 & .0070 \\

(c) &  & \cmark &  &
.9397 & .9119 & .9794 & .0090 &
.9354 & .8853 & .9710 & .0052 \\

(d) &  &  & \cmark &
.9326 & .9058 & .9778 & .0107 &
.9310 & .8797 & .9758 & .0060 \\

(e) & \cmark & \cmark &  &
.9530 & .9273 & .9865 & .0062 &
.9421 & .8972 & .9831 & .0056 \\

(f) & \cmark &  & \cmark &
.9474 & .9275 & .9860 & .0072 &
.9329 & .8862 & .9782 & .0052 \\

(g) &  & \cmark & \cmark &
.9434 & .9220 & .9849 & .0078 &
.9404 & .8956 & .9822 & .0050 \\

\midrule
Ours& \cmark & \cmark & \cmark &
\textbf{.9582} & \textbf{.9383} & \textbf{.9924} & \textbf{.0058} &
\textbf{.9449} & \textbf{.9009} & \textbf{.9838} & \textbf{.0048} \\
\bottomrule
\end{tabular}
}
\end{table*}

\begin{figure*}[!t]
  \centering
  \includegraphics[width=\linewidth]{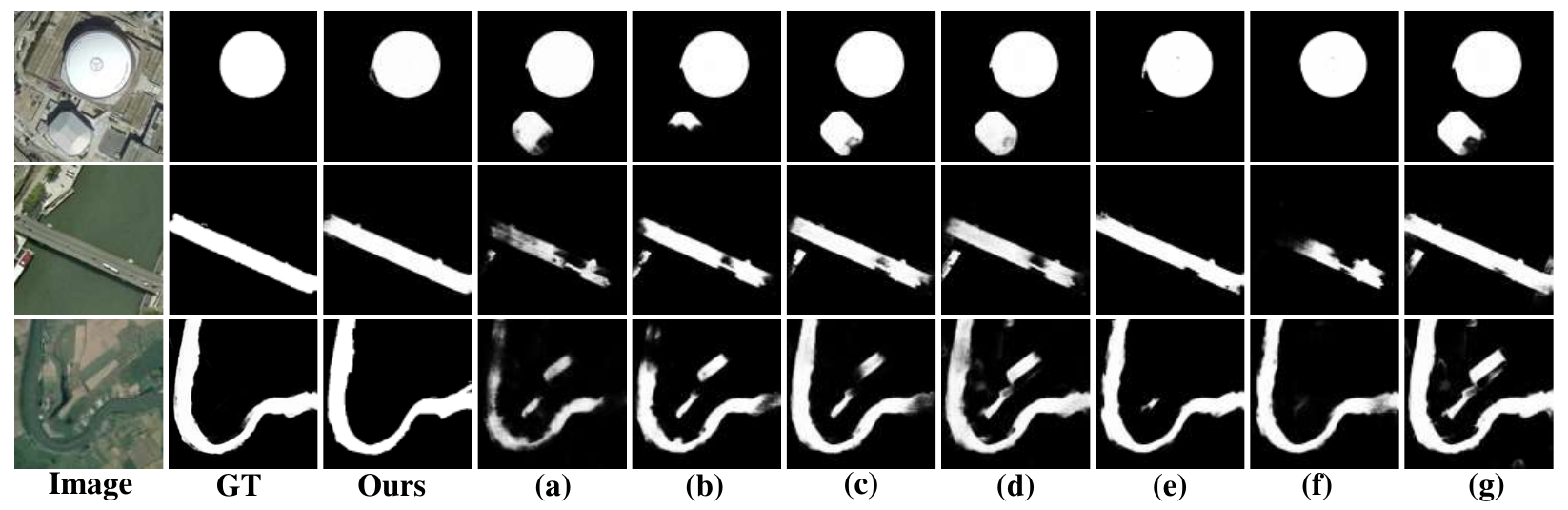}
  \caption{Visual comparison of qualitative results produced by models with different ablation settings. The complete model achieves the best performance in terms of boundary delineation and accurate object identification.}
  \label{fig:Ablation}
\end{figure*}

\textbf{Qualitative Comparisons.}
Figure \ref{fig:Visualization} presents a qualitative comparison between our model and six of the latest  state-of-the-art methods from recent years on the ORSSD and EORSSD datasets. We selected multiple representative ORSIs scenarios, including large objects, small objects, elongated structures, and low-contrast objects. Compared with other methods shown in the figure, our model demonstrates superior performance in saliency detection under these complex conditions.
For example, in the low-contrast scene in the fourth row, only our model is able to accurately segment the complete contour of the ship, whereas other methods fail to correctly identify the low-contrast ship and exhibit varying degrees of segmentation errors, indicating their limited adaptability to such challenging conditions.
This highlights our model’s strong capability in capturing subtle local variations and performing precise boundary localization. 
Similarly, in the fifth row, which involves elongated structures detection, only our model is able to detect the complete contour of the target, whereas the other methods are almost unable to effectively recognize it. This result clearly demonstrates that the proposed model, while fully leveraging the long-range dependency modeling capability of the state space architecture, also exhibits excellent fine-grained boundary localization and structural perception abilities.
Moreover, in the small objects scenario shown in the last row, only our method able to accurately detect all salient objects, whereas the other methods generally suffer from missed detections. This phenomenon further indicates that our model possesses stronger target discrimination capability and scale robustness in complex multi-scale scenes.
In summary, the qualitative results in Figure \ref{fig:Visualization} comprehensively demonstrate the robustness and accuracy of our proposed model in salient object detection across various complex ORSIs scenes. Our approach not only ensures effective global context modeling but also excels at extracting critical local features, enabling the model to learn saliency information more effectively.

\subsection{Ablation Studies}
\textbf{Effectiveness of Key Components.}
We evaluate the effectiveness of the proposed MLCM, GPM, and DAGS through a series of ablation experiments on the ORSSD and EORSSD datasets, as summarized in Table~\ref{tab:module_ablation}. Starting from a baseline model without any proposed modules, introducing each component individually leads to consistent performance improvements. Specifically, MLCM enhances global structural perception via multi-level context aggregation, GPM yields more pronounced gains by reinforcing global contextual priors and suppressing background interference, while DAGS further improves boundary completeness and structural coherence through resolution-aware sequence modeling. Combining any two modules consistently outperforms their single-module counterparts, indicating strong complementarity among them. The best performance is achieved when all three modules are jointly employed, yielding the highest scores across all evaluation metrics on both datasets. The qualitative results in Figure \ref{fig:Ablation} further support these findings: the baseline model tends to produce incomplete predictions with fragmented boundaries and background confusion, while the full model generates more coherent object regions with sharper boundaries and fewer false positives, especially in challenging scenarios with complex backgrounds and elongated structures. These results demonstrate that MLCM, GPM, and DAGS are highly complementary and collectively contribute to robust and accurate salient object detection.

\textbf{DAGS with other scanning strategies.}
Table~\ref{tab:scan_ablation} compares the proposed DAGS with several representative scanning strategies on the EORSSD dataset. As shown, DAGS consistently outperforms Cross Scan\cite{VMamba_2024}, Diagonal Scan\cite{xu2025direction}, Hilbert Scan\cite{hibert2024}, and Local Scan\cite{localmamba} across all metrics, achieving the highest $F_{\beta}^{max}$, $E_{\xi}^{max}$, and $S_{\alpha}$, while yielding the lowest MAE. These results demonstrate that the proposed DAGS is more effective in capturing spatial dependencies and structural information, thereby validating the superiority of our scanning strategy.
\begin{table}[t]
\centering
\caption{Ablation for the impact of different scanning.}
\label{tab:scan_ablation}
\resizebox{\linewidth}{!}{
\begin{tabular}{c|c|cccc}
\toprule
\multirow{2}{*}{ID} & \multirow{2}{*}{Scanning Method} & \multicolumn{4}{c}{EORSSD} \\
\cmidrule(lr){3-6}
 &  & $F_{\beta}^{max} \uparrow$ & $E_{\xi}^{max} \uparrow$ & $S_{\alpha} \uparrow$ & $MAE\downarrow$ \\
\midrule
1 & Cross Scan~\cite{VMamba_2024}        & .9421 & .8972 & .9831 & .0056 \\
2 & Diagonal Scan~\cite{xu2025direction}    & .9391 & .8917 & .9782 & .0064 \\
3 & Hilbert Scan~\cite{hibert2024}  & .9384 & .8906 & .9755 & .0057 \\
4 & Local Scan~\cite{localmamba}  & .9407 & .8947 & .9812 & .0059 \\
5  & \textbf{DAGS (Ours)}             & \textbf{.9449} & \textbf{.9009} & \textbf{.9838} & \textbf{.0048} \\
\bottomrule
\end{tabular}
}
\end{table}

\section{Conclusion}
In this paper, we propose ASCNet, a novel network for ORSI-SOD, designed to address the challenges of complex backgrounds, large-scale variations, and intricate spatial structures in optical remote sensing images.
Specifically, we introduce a MLCM to enable interactive fusion of multi-scale features across different representation levels. By leveraging topology-aware attention to model cross-spatial and cross-regional relationships, MLCM effectively aggregates contextual information from multiple semantic levels and spatial regions, resulting in more discriminative saliency representations with improved robustness and generalization.
To enhance local region modeling in sequence-based visual representations, we further integrate an APVSS block into the decoder, which consists of GPM and DAGS. The GPM employs learnable global semantic tokens to semantically condition local patch features, strengthening local structural perception while maintaining global semantic consistency. Meanwhile, DAGS adopts a resolution-aware, block-wise scanning strategy on multi-scale feature maps, adjusting scan granularity by resolution and dynamically reweighting directional scans to balance local detail modeling and global contextual awareness during decoding.
Extensive experiments on multiple ORSI-SOD benchmarks demonstrate that ASCNet consistently achieves state-of-the-art performance, validating the effectiveness of the proposed modules and overall architecture.

\bibliographystyle{elsarticle-num}
\bibliography{main}





\end{document}